\documentclass{article}

\usepackage{graphicx}
\usepackage{algorithm,algorithmic}
\usepackage{amsmath}
\usepackage{amssymb}
\usepackage{amsthm}
\usepackage{subfigure}

\newcommand{\eg}{\textit{e.g.} }
\newcommand{\ie}{\textit{i.e.} }

\begin{document}

\title{Active Testing for Face Detection and Localization}
\author{Raphael Sznitman \\ \textit{Department of Computer Science} \\ \textit{Johns Hopkins University} \and Bruno Jedynak \\ \textit{Department of Applied Mathematics and Statistics} \\ \textit{Johns Hopkins University}}

\date{}

\maketitle

\begin{abstract}
We provide a novel search technique, which uses a hierarchical model
and a mutual information gain heuristic to efficiently prune the
search space when localizing faces in images. We show exponential
gains in computation over traditional sliding window approaches,
while keeping similar performance levels.
\end{abstract}

\section{Introduction}
\label{sec:intro}

In recent years, face detection algorithms have
provided extremely accurate methods to localize faces in images.
Typically, these have involved the use of a strong classifier, which
estimates the presence of a face given a particular subwindow of the
image. Successful classifiers have used Boosted Cascades
\cite{viola04rapid, schneiderman04, li04floatboost, gao08}, Neural
Networks \cite{rowley96, lecun07, delakis04} and SVM's
\cite{freund97,Shah07} among others.

In order to localize faces, the aforementioned algorithms have
relied on a sliding window approach. The idea is to inspect the
entire image by sequentially observing each and every location a
face may be in by using a classifier. In most face detection
algorithms \cite{viola04rapid, li04floatboost, lecun07, gao08}, this
involves inspecting all pixels of the image for faces, at all
possible face sizes. This exhaustive search, however, is
computationally expensive and in general not scalable to large
images. For example, for real-time face detection using modern
cameras (4000 $\times$ 3000 pixels per image), more than \textit{100
million} evaluations are required, making it hopeless on any
standard computer.

To overcome this problem, previous works in object and face
localization have simply reduced the pose space by allowing only a
coarse grid of possible locations \cite{triggs05, rowley96,
viola04rapid}. An elegant improvement to object detection was
proposed in \cite{schneiderman04} where ``feature-centric"
evaluation are performed, as opposed to ``window-centric", allowing
previous computation to be reused. Such a method however relies on
strong knowledge of the classifier used. More recently, a globally
optimal branch-and-bound subwindow search method for objects in
images was proposed \cite{Lampert08} and extended to videos
\cite{yuan09}. Here, the classifier and the feature space used to
locate the object are dependent on a single robust feature (\eg SIFT
\cite{lowe04}), making it difficult to use in the context of faces.

In this paper, we propose a novel search strategy, which can be
combined with any face classifier, in order to significantly reduce
the computational cost involved with searching the entire
space. The design principle is as follows: We assume that a
\textit{perfect} face classifier is available, \ie one which always
provides the correct answer. In practice however, such a classifier
does not exist and an accurate one (as in
\cite{viola04rapid, li04floatboost, lecun07, gao08}) will be used instead. Our
goal is then to reduce the total number of classifier evaluations
required to detect and locate faces in images, while still providing
similar performance levels when compared with an exhaustive search.

A proposed strategy for computational shape recognition
\cite{geman93}, argues that the task of visually recognizing an
object can be accomplished by querying the image in a sequential and
adaptive way. In general, this can be regarded as a coarse-to-fine
approach to perception \cite{amit99, viola04rapid, fleuret01,
fleuret08}. This ``twenty questions" approach can be described as
follows: there is a fact to be verified, \eg ``is there a face in
the field of view", and each query, which consists of evaluating a
particular function of the image, is chosen to maximally reduce the
expected uncertainty about this fact. In the context of computer
vision, such approaches have led to two different types of search
algorithms: offline and online. In the offline versions, the ``where
to look next" strategy is computed once and for all, anticipating
all possible queries. It has led to efficient algorithms for symbol
recognition \cite{amit99}, face \cite{fleuret01} and cat
\cite{fleuret08} detection. In the online version, the strategy is
computed sequentially, as information is gathered. It has led to a
road tracking algorithm \cite{geman93,geman96}: this approach is
known as \textit{Active Testing}.

In this paper, we extend the active testing framework in order to do
fast face detection and localization. We provide a way to ask
questions that are general and specific with regard to the face
pose, and span different feature spaces. Similarly to the ``twenty
questions'' game, questions such as ``is the object at this location
with this size?'' are asked by means of an accurate face classifier
\cite{viola04rapid,Shah07,lecun07,gao08}, independently of what
features are used to guide the search. We show here that this
approach provides a coherent framework, with few parameters to
choose or tune, which significantly reduces the number of classifier
evaluations necessary to localize faces. Comparison of our method
with state-of-the-art face detection algorithms, and the traditional
sliding window approach, indicate that our framework reduces, by
several orders of magnitude, the number of classifier evaluation
needed while maintaining similar accuracy levels on localization and
detection tasks. Even though this paper specifically focuses on
frontal faces, this approach can be extended to faces in general \cite{torre07,jiang08,martinez10,prince09,schiffer08}, other object
categories \cite{pascal-voc-2009} and to most classifiers in the machine learning
literature.

The remainder of this paper is organized as follows: in section
\ref{sec:model}, the general framework of our method is presented
along with implementation details. Section \ref{sec:experiment_loc}
describes localization experiments, and in section
\ref{sec:exp_multi} we compare the performance with state-of-the-art
methods on a detection and localization task. Concluding remarks are
provided in section \ref{sec:conclusion}.

\begin{figure*}
\centering \subfigure[ ] { \label{fig:tree}
\includegraphics[width=0.3\textwidth]{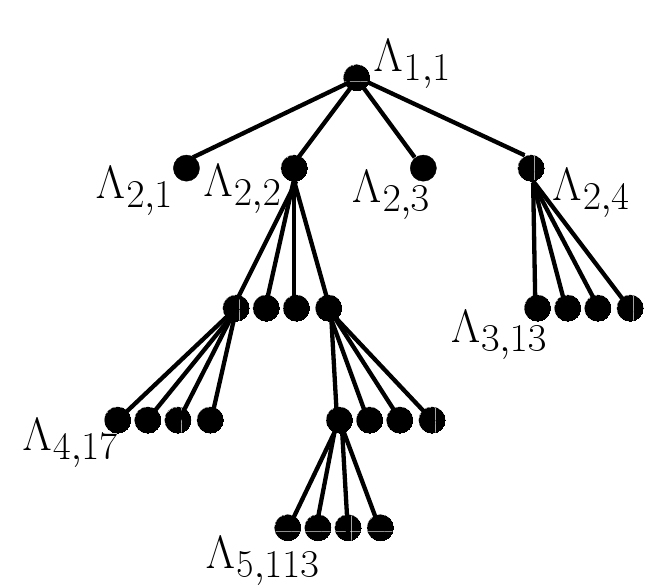}}
\subfigure[ ]{\label{fig:windows}
\includegraphics[width=0.3\textwidth]{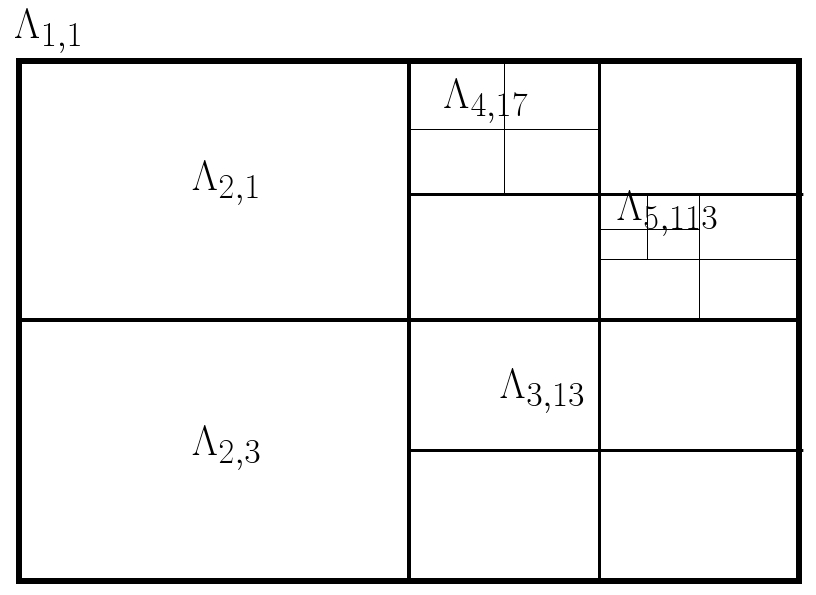}}
\subfigure[ ]{\label{fig:queries}
\includegraphics[width=0.3\textwidth]{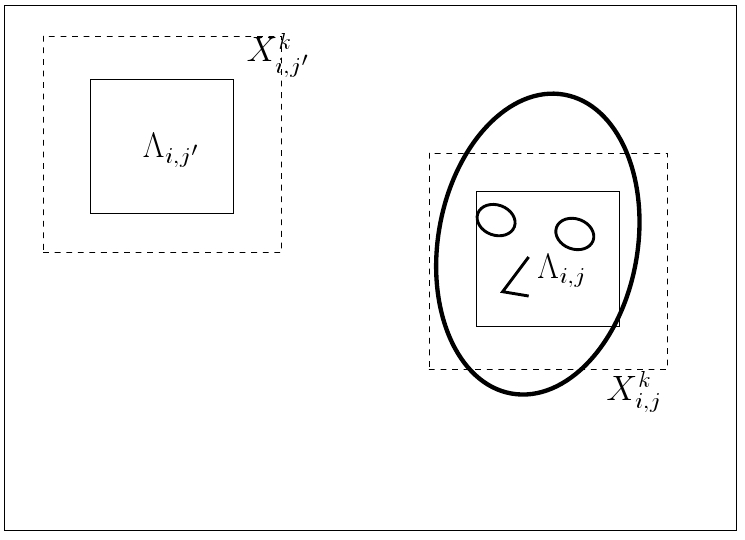}}
\caption{ \small Each node in the tree (a) corresponds
to subwindow in the image (b). The root of the tree,
 $\Lambda_{1,1}$, represents the entire image space and has four children ($\Lambda_{2,1}, \Lambda_{2,2},
 \Lambda_{2,3}, \Lambda_{2,4}$). (c) Example Query: Here, the face center is, $Y=l \in \Lambda_{i,j}$.
 The query $X^k_{i,j}$, counts the proportion of edges in a window twice the size of $\Lambda_{i,j}$, centered on
 $\Lambda_{i,j}$. $k$ indicates that we count the proportion of edges on a surface twice the size of the subwindow
 $\Lambda_{i,j}$, while $\{i,j\}$ provides the pose subset in $\Lambda$.
} \label{fig:model}
\end{figure*}

\normalsize

\section{Active Testing}
\label{sec:model}

The goal set forth is to detect and localize a single frontal face
of unknown size, which may or may not be present in the image. We
define the pose of a face, as the pixel location of the face center
and a face scale. That is, we treat localization as placing a
bounding box around a face. In section \ref{sec:exp_multi}, we
detail how this can be extended to searching for multiple faces.

Active Testing (AT) can be regarded as a search algorithm which uses
an information gain heuristic in order to find regions of the search
space which appear promising. The region which is to be observed
next is determined as information is gathered, and thus can be
viewed as an \textit{online} variation of the ``twenty questions''
game. The general approach is as follows: we are looking for a face
in an image, and are provided with a set of questions which help us
determine where the face is located. Questions are answered with
some uncertainty, reducing the search space and eventually leading
to the face pose.

In addition, it is also assumed that a special question regarding
the exact face pose is available. This question is treated as an
``Oracle'', always providing a \textit{perfect} answer when queried
but is computationally expensive relative to other questions.
Querying the oracle at every location would provide the face pose
but is expensive and inefficient as certain questions are more
informative than others and help reduce the search space faster.
Consequently, a sub-goal is to determine face pose with as few
questions as possible.

\subsection{Model and Algorithm}
\label{sec:mod_phil}

Let $Y=(L,S)$ be a discrete
random variable defining the face pose; where $L$ is the location of
the face center (\ie pixel coordinates), and $S$ is the face scale,
such that $S$ can take values $\{1,\ldots, M\}$ corresponding to $M$
face size intervals. Additionally, $Y$ can take one extra value when
the face is not in the image. Let

$$\Lambda = \{ \Lambda_{i,j}, i = 1,\ldots,D, j = 1,\ldots,4^{i-1} \}$$

be a quadtree of finite size, which decomposes the image space; $i$
indexes the level in the tree and $j$ designates the cell at that
level (see figure \ref{fig:model}(a)). Every leaf is associated with a
pixel in the image and each non-terminal node corresponds to a
unique subwindow in the image, representing a subset of poses
(figure \ref{fig:model}(b)). When no face is present in the image
then $Y \in \bar{\Lambda}_{1,1}$, where $\bar{\Lambda}_{1,1}$
denotes the complement of $\Lambda_{1,1}$.

We are interested in refining the estimate of where the face is
located iteratively and hence denote $\pi_t$ as the probability
density of $Y$ at iteration step $t$. Let $u_{i,j,s} = P(L \in
\Lambda_{i,j}, S = s), \Lambda_{i,j} \subset \Lambda, s \in
\{1,\ldots,M\}$. By construction, calculating $u_{i,j,s}$ can be
achieved by summing the probability of $\Lambda_{i,j}$'s children.
Clearly, $u_{1,1,s} = u_{2,1,s}+ u_{2,2,s} + u_{2,3,s}+u_{2,4,s}$
and similarly for any other $u_{i,j,s}$. For any node, we also
denote $u_{i,j} = \pi(\Lambda_{i,j})= \sum^M_{s=1} u_{i,j,s}$. Let
$\mathcal{X} = \{ \mathbf{X}^1,\ldots,\mathbf{X}^K \}$ be a set of
question families, such that for each family $k$, $\mathbf{X^k} = \{
X^k_{i,j}, i = 1,\ldots,D, j = 1, \ldots,4^{i-1}\}$, where
$X^k_{i,j}$ is a query from family $k$, about the pose subset
$\Lambda_{i,j}$.

The generic AT algorithm (algorithm 1) can then be seen as the
following: to begin, $\pi_0$ and the first query are initialized
(lines 1 to 2). Three operations are then repeated: the response is
observed (line 4); the belief of the location of $Y$ is updated
using the latest observation (line 5); a new query is chosen for the
next iteration (line 6). The iteration is stopped when a terminating
criteria is achieved (line 7). Each line is explained in detail in
the following sections.

\begin{algorithm}
\caption{Active Testing (AT)}
\begin{algorithmic}[1]
\STATE Initialize: $i \leftarrow 1, j \leftarrow 1, k \leftarrow 1,t
\leftarrow 0$ \STATE Initialize: $\pi_0(\Lambda_{1,1}) =
\pi_0(\bar{\Lambda}_{1,1}) = \frac{1}{2}$ \REPEAT
 \STATE Compute the test $x=X^k_{i,j}$
 \STATE Compute $\pi_{t+1}$ using $\pi_t$ and $x$
 \STATE Choose the next subwindow and test:
  $$ \{i,j,k\} = \arg \max_{i',j',k'} I(Y;X^{k'}_{i',j'})$$
\UNTIL $H(\pi_{t+1}) > 1-\epsilon$ and/or $t < \gamma$.
\end{algorithmic}
\label{alg:at}
\end{algorithm}

\subsection{Queries}
\label{sec:mod_queries}

The AT algorithm requires a set of query families, $\mathcal{X} = \{
\mathbf{X}^1,\ldots,\mathbf{X}^K \}$, to be specified. Each query
family, $\mathbf{X}^k$, consists of evaluating a specific type of
image functional indexed by $k$. Members of a family, $\mathbf{X^k}
= \{ X^k_{i,j}, i = 1,\ldots,D, j = 1,\ldots,4^{i-1}\}$, are indexed
by a pose index in $\Lambda$ (as in \cite{fleuret08}). That is,
$X^k_{i,j}$ is an image functional, where $k$ defines a particular
computation and $\{i,j\}$ specifies the pose subset. Note that these
queries are generic and need not be binary. Example queries can be
seen in Figure \ref{fig:model}(c).

In addition, \textit{perfect} tests - which precisely predict the
presence of a face by using a classifier - are included in
$\mathcal{X}$. When this test is used at a specific pose, either the
classifier responds positively and the face is deemed found, or
conversely, the response is negative and the face is assumed not to
be at this pose. That is, we assume no uncertainty with regard to
the response of this classifier.

In order to specify the joint distribution between the face pose $Y$
and queries $\mathcal{X}$, we make the following heuristic
assumptions:

\textbf{Conditional Independence}
\begin{eqnarray}
\label{eq:condInd}
P\left(\{X^k_{i,j}=x\}, i = 1\ldots D,j=1 \ldots 4^{i-1}, k=1 \ldots K |Y=(l,s)\right) & & \nonumber \\
 = \prod_{i,j,k} P\left(X^k_{i,j}=x |Y=(l,s)\right) & &
\end{eqnarray}

 \textbf{Homogeneity}

\begin{equation}
\label{eq:homogeneity} P\left(X^k_{i,j}=x|Y=(l,s)\right) = \left\{
\begin{array}{ll}
f^{k}_{s}(x; i) & \textrm{ if } l \in \Lambda_{i,j} \\
f^{k}_{0}(x; i) & \textrm{ otherwise }
\end{array}
\right.
\end{equation}

Here $f^{k}_{s}$ characterizes the ``response" to the query
$X_{i,j}^k$ when the center of the face is within $\Lambda_{i,j}$
with size $s$. Similarly, $f^{k}_{0}$ is the ``response'' when the
center is not in $\Lambda_{i,j}$. Additionally, even though $KN$
queries are specified, where $N$ is the number of nodes in
$\Lambda$, the number of densities needed is only $KD$. That is, for
each test family, only one density per level of $\Lambda$ needs to
be specified. This is why $f_s^k(\cdot, i)$ is only indexed by $i$.

Note that these assumptions are a simple way to make the problem
tractable: for example, the conditional independence of queries
given the location of the object $Y$ assumption is clearly a
simplification as the same pixel values are used to compute many
queries at different levels of $\Lambda$. Similarly, the actual
responses to tests might in fact depend on the precise location of
the face within $\Lambda_{i,j}$. The homogeneity assumption
simplifies the response model by assuming a single model for all
cases. Even when using these assumptions however, the experiments
conducted here (sections \ref{sec:experiment_loc} and
\ref{sec:exp_multi}) indicate that these simplifications provide a good
way to solve the problem at hand. In addition, this model should be taken into account when choosing
queries to use: similarly to a Naive Bayes model, queries should be
individually informative.

\subsection{Belief Update}
\label{sec:mod_update}

Once an observation has been made, the new distribution of the face
location $Y$ must be calculated (line 5 of AT). At initialization
(line $1$ of AT), $\pi_0(\Lambda_{1,1}) = \pi_0(\bar\Lambda_{1,1}) =
\frac{1}{2}$, indicating that a face is believed to be in the image
with probability $1/2$. Note that the probability
$\pi_0(\Lambda_{1,1})$ is uniformly distributed within
$\Lambda_{1,1}$ by construction. Given $\pi_t$ and the query
response $X^k_{i,j} = x$ at time step $t$, the updated distribution
$\pi_{t+1}$ can then be calculated by using Bayes formula

\small
\begin{equation}
\label{eq:bayes} \pi_{t+1}(l,s) = \frac{P\left(X^k_{i,j} = x|Y
=(l,s)\right)\pi_t(l,s)}{\sum_{s'} \int_{l'} P\left(X^k_{i,j} =
x|Y=(l',s')\right)\pi_t(l',s')dl'}
\end{equation}
\normalsize

\noindent Using assumptions \ref{eq:condInd} and
\ref{eq:homogeneity} then

\begin{equation}
P\left(X^k_{i,j} = x|Y =(l,s)\right) = f^k_0(x,i)1{\hskip
-2.5pt}\hbox{I}_{\bar{\Lambda}_{i,j}}(l) + f^k_s(x,i)1{\hskip
-2.5pt}\hbox{I}_{\Lambda_{i,j}}(l)
\end{equation}

\noindent Let us now define the likelihood ratio as
\begin{equation}
r(x,s) = \frac{f_s^k(x,i)}{ f_0^k(x,i)}, s=1\ldots M
\end{equation}

\noindent then equation \ref{eq:bayes} can be written as,

\begin{equation}
\pi_{t+1}(l,s) = \frac{1}{\mathcal{Z}(x)}\left(1{\hskip
-2.5pt}\hbox{I}_{\bar{\Lambda}_{i,j}}(l) + 1{\hskip
-2.5pt}\hbox{I}_{ \Lambda_{i,j}}(l)r(x,s)\right)\pi_t(l,s)
\label{eq:faceUpdateEq}
\end{equation}

\noindent
where $\mathcal{Z}(x) $ is the normalizing constant,
\begin{equation}
  \mathcal{Z}(x) = \pi_t(\bar{\Lambda}_{i,j})+ \sum_{s=1}^M r(x,s)\pi_t(\Lambda_{i,j})
\end{equation}
Note that the evolution from $\pi_{t}$ to $\pi_{t+1}$ only
relies on $r(x)$ and allows for probability mass to be shifted onto
or away from $\Lambda_{i,j}$, depending on the response of
$X^k_{i,j}$.

In order to reduce the number of nodes to update, only a subtree is
maintained, where only nodes which have probability greater than
some threshold $\tau$ are included. By construction of $\Lambda$,
parent nodes have probability equal to the sum of their children,
hence any node which has probability larger than $\tau$ also has
parent with probability greater than $\tau$. This guarantees that
applying this threshold forms a subtree within $\Lambda$ containing
$\Lambda_{1,1}$. This approximation of $\pi_t$ allows for a compact
representation of the distribution.

\subsection{Query Selection}
\label{sec:mod_phi} We choose to select the next query by maximizing
the mutual information gain between $Y$ and the possible queries
$X^k_{i,j}$ (line 6 of AT). This can be written as

\begin{equation}
 I(Y;X^k_{i,j}) = H(X^k_{i,j}) - H(X^k_{i,j}|Y)
\end{equation}

\noindent
where,

\begin{equation}
H(X^k_{i,j}) = h \left(\sum_{s=0}^M u_{i,j,s} f^k_s(\cdot) \right)
\end{equation}

\noindent here, $h(f)$ is the differential Shannon entropy of the
density $f$. We simplify this expression by substituting $h(f)$ with
the Gini Index \cite{elemStat}. The mutual information then becomes

\begin{equation}
\label{eq:phiFaceScale}
I(Y;X^k_{i,j}) = \sum_{s=0}^M \sum_{m>s}^M u_{i,j,s} u_{i,j,m} \int(f^k_s - f^k_m)^2
\end{equation}

\noindent where $u_{i,j,0} = 1 - u_{i,j}$. Note that the term $\int
(f^{k}_{s} - f^{k}_{m})^2$ is the Euclidean distance between the
densities $f^{k}_{s}$ and $f^{k}_{m}$, and only needs to be computed
once and then stored for fast evaluation.

Since we are interested in choosing both the region
$\Lambda_{i,j}\in \Lambda$ and a query family $k$ which maximizes
the information gain, one can simply evaluate $I(Y;X^k_{i,j})$ for
all possible values of the triple $(i,j,k)$ and select the
parameters providing the largest gain. However, as described in
section \ref{sec:mod_update}, only a small subset of poses is ever
considered at any iteration. For example, nodes which have little
probability will surely only provide a small information gain.
Consequently, we only need to evaluate equation
(\ref{eq:phiFaceScale}) for the explicitly maintained subtree
(Figure \ref{fig:model}(a)). Additionally once a query has been chosen,
it is removed from the set of possible queries, further reducing the
amount of computation.

\subsection{Terminating Criteria}
\label{sec:mod_termin}

At line 7 of the AT algorithm, two terminating criteria are
presented: (i) the algorithm runs until the entropy of $\pi$, $H(
\pi)$, is very high, (ii) the algorithm iterates for a fixed number
of steps, $\gamma$. In the first case, running until the entropy is
high corresponds to two possible outcomes: either a face has been
found and most of the probability mass is at a single leaf of
$\Lambda$ or most of the mass is outside the image,
$\bar{\Lambda}_{1,1}$ and no face is believed to be present in the
image. In general, the choice of which criteria to use ((i), (ii) or
both) is for the user to decide. Sections \ref{sec:experiment_loc}
and \ref{sec:exp_multi} show the behavior of these scenarios.

In addition, for all cases, the total number of queries is bounded
by the size of the tree and the number of query families. As the
algorithm iterates and the classifier is queried, the number of
poses with strictly positive probability decreases. This provides a
guarantee that, in the worst case, the face will be found after
having observed all the poses.

\begin{figure*}
\centering
\includegraphics[width=0.9\textwidth]{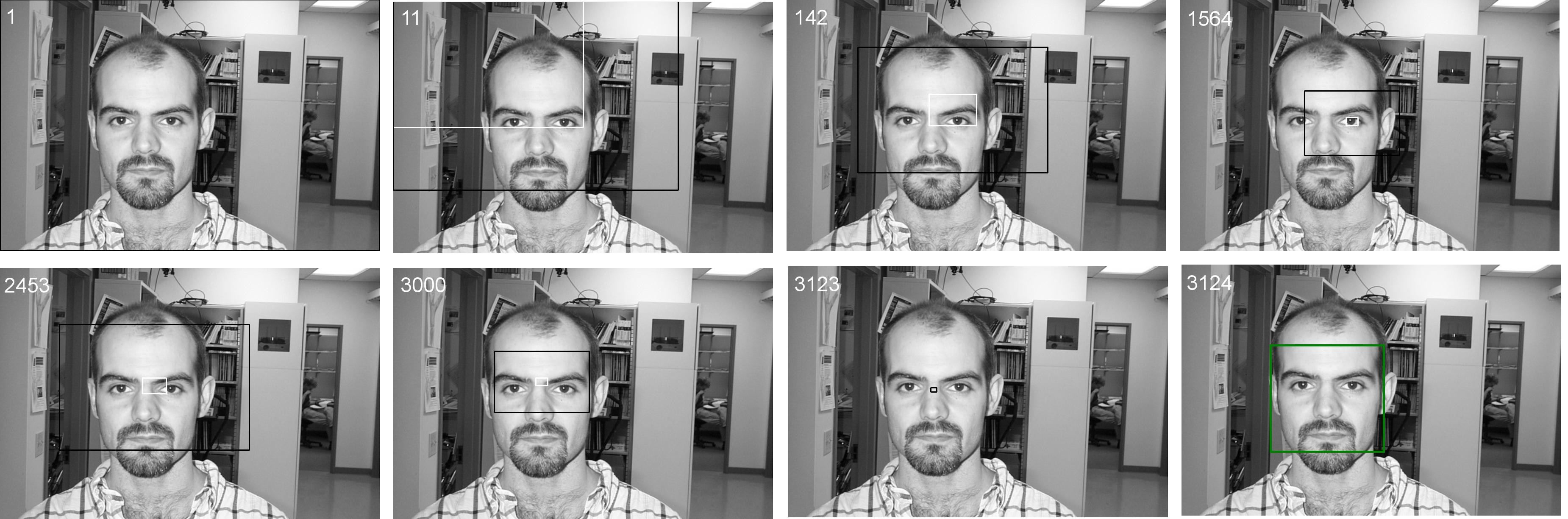}
\caption{ \small Sequence of queries posed by the Active Testing
algorithm on a test image from the Caltech Frontal Face Dataset. In
each image, a test $X^k_{i,j}$ is computed: white boxes show the
pose, $\Lambda_{i,j}$, queried while black boxes show the subimage
queried. The number indicated in the top left of each image is the
iteration number of the AT algorithm. In image 3123, the Boosted
Cascade is evaluated and a face is found at a given scale (green
box). } \label{fig:seq}
\end{figure*}
\normalsize

\subsection{Implementation}
\label{sec:face_imple}

We now provide some implementation details and give a more in depth
algorithm for updating $\pi$ (see algorithm \ref{alg:update}) and
choosing queries.

Before the AT algorithm begins, all features necessary to evaluate
queries from $\mathcal{X}$ for a given image are computed and stored
in the form of an integral image making the evaluation of a query
$O(1)$ operations (similar to \cite{Lampert08}). This is
particularly efficient since queries $X^k_{i,j}$ compute nested
subwindows.

In order to form and maintain the subtree of $\Lambda$ (line 7),
only nodes which are above a threshold ($\tau = 0.001$) are
explicitly stored. To do this, we construct $\Lambda$ as a quadtree,
and maintain a frontier set $\mathcal{F}$. $\mathcal{F}$ consists of
any node $\Lambda_{i,j}$ with $u_{i,j} > \tau$ and with all children
having $u_{i+1,j'} < \tau$. Applying this rule at each iteration
ensures that the maintained subtree is relevant to where the face is
believed to be located. Additionally, since the probability
associated at any node in the tree is equal to the sum of its
children, we only need to update nodes in $\mathcal{F}$, and recurse
through the tree to update the remaining nodes in $\Lambda$.

After having computed the query $X_{i,j}^k$, updating any node
$\Lambda_{i',j'} \in \mathcal{F}$ is simple: if $\Lambda_{i',j'} \in
\Lambda_{i,j}$, then $u_{i',j'} = r(y)u_{i',j'}/\mathcal{Z}$,
otherwise $u_{i',j'} = u_{i',j'}/\mathcal{Z}$. Doing so updates
$\pi$ as described in equation (\ref{eq:faceUpdateEq}) in an
efficient way. In addition, at any point in the updating of $\pi$,
the next best query, $S$, seen so far is maintained. The denominator
$\mathcal{Z}$ is calculated once and for all, and used to calculate
equation \ref{eq:phiFaceScale} when each node is visited. Only the
best score is kept, and ultimately chosen for the following
iteration of the AT algorithm. That is, we compute equations
(\ref{eq:faceUpdateEq}) and (\ref{eq:phiFaceScale}) one after the
other, requiring only one pass through the subtree per iteration.

\begin{algorithm}
\caption{Update($\Lambda_{i',j'},\Lambda_{i,j}, x, S, \mathcal{F}$)}
\begin{algorithmic}[1]
 \IF {$\Lambda_{i',j'} \in \mathcal{F}$}
  \IF {$\Lambda_{i',j'} \subset \Lambda_{i,j}$}
   \STATE $u_{i',j'} \leftarrow r(x)u_{i',j'}/\mathcal{Z}$
  \ELSE
   \STATE $u_{i',j'} \leftarrow u_{i',j'}/\mathcal{Z}$
  \ENDIF
  \STATE Maintain $\mathcal{F}$
 \ELSE
  \FOR{Each child, $\Lambda_{i'+1,j''}$, of $\Lambda_{i',j'}$ }
   \STATE Update($\Lambda_{i'+1,j''},\Lambda_{i,j},x,S,\mathcal{F}$)
   \ENDFOR
   \STATE $u_{i',j'} \leftarrow \sum_{j''} u_{i'+1,j''}$
 \ENDIF
 \STATE $S = \max \left(S, \max_k I(Y;X^k_{i',j'}) \right)$
\end{algorithmic}
\label{alg:update}
\end{algorithm}

\section{Face Localization}
\label{sec:experiment_loc}

To demonstrate that this framework can be used to significantly
reduce the number of classifier evaluations required when searching
for a face in an image, we begin by evaluating the AT algorithm on a
pure localization task (as done in \cite{Lampert08}). In the
following set of experiments, each image contains exactly one face.
We describe in section \ref{sec:face_queries} the queries used to
localize faces. In section \ref{sec:exp_single} we show how AT
performs in terms of time, number of classifier evaluations and
accuracy.

We perform the following experiments on the Caltech Frontal Face
dataset \cite{weber99}, which consists of 450 images (896 $\times$
592 pixels), each containing exactly one of 27 different faces in
variously cluttered environments and illuminations. Face sizes range
from approximately 100 to 300 pixels in width. We choose $M = 4$
possible face size intervals ($[100,150], [150,200], [200,250],
[250,300]$). All experiments are conducted on a 2.0 Gigahertz
machine.

\subsection{Face Queries}
\label{sec:face_queries}

To locate faces, we first specify the following set of test
families, $\mathcal{X} = \{\mathbf{X}^1, \ldots, \mathbf{X}^K \}$
and their associated distributions ($f_s^k, f_0^k$). In the
following experiments, $K = 30$.

The first family of tests, $\mathbf{X}^1$, calculates the proportion
of edge pixels (defined and computed as in \cite{amit99} by means of
an edge oriented integral image) in a window associated with the
pose $\Lambda_{i,j}$. That is, $X^1_{1,1}$ is the proportion of
pixels which are edges within $\Lambda_{1,1}$ and similarly for all
$\Lambda_{i,j}$. Test families $\mathbf{X}^2$ to $\mathbf{X}^5$ are
similar to $\mathbf{X}^1$, in that they compute the proportion of
edge pixels in a window centered on $\Lambda_{i,j}$, but of larger
size, by a factor $F = \{ 2,3,4,5 \}$ (see figure \ref{fig:model}(c)).
Note that this factor is different from the scale $S$. Using these
pose-indexed tests provides a way to test arbitrarily large regions,
even when $\Lambda_{i,j}$ is a small subwindow. These tests also
allow for overlapping $\Lambda_{i,j}$ regions and more precise
estimation of the face scale.

Families $\mathbf{X}^6$ to $\mathbf{X}^9$ are similar to
$\mathbf{X}^1$ but compute the proportion of edge pixels in a
particular direction (four possible directions). Similarly to
families $\mathbf{X}^2$ to $\mathbf{X}^5$, families
$\mathbf{X}^{10}$ to $\mathbf{X}^{25}$ allow for a scale factor for
tests in a particular direction (4 directions $\times$ 4 factors).
Using integral images allows for computation of these tests with
only 4 additions, making them very efficient.

We choose to model all the $f_s^k$ for $s \in \{0,...,M\}$ using
Beta distributions. The Beta family permits to model a wide range of
smooth distributions over the interval $[0,1]$ with only two
parameters. The parameters of each distribution are determined
offline from a small training dataset where the face location and
scale is known (more details are given in Section
\ref{sec:exp_train}).

Finally, families $\{\mathbf{X}^{26},\ldots,\mathbf{X}^{30}\}$ are
the \textit{perfect} tests and involve testing for a face using a
Boosted Cascade (BC). Each family specifies testing for a face at all
scales within a given interval ($s \in \{1,...,M \}$). For each
interval, we test for face sizes in increments of 10\% of the
smallest face size (total of 13 face sizes in the range
$[100,300]$). In terms of operations, evaluating this test requires
on average 56 additions, 1 multiplication and 1 comparison, per face
size, making it significantly more costly than other queries. Since
the BC is only informative when the pose is very
specific, we restrict this test to leaves in $\Lambda$. These
BCs are trained and provided by OpenCv \cite{opencv},
but modified to restrict testing to specific regions and face sizes.
Even though better classifiers have recently been developed, we
choose this one as it is publicly available and widely used.

\begin{figure*}
\centering \subfigure[
]{\label{fig:loc_roc}\includegraphics[width=0.3\textwidth]{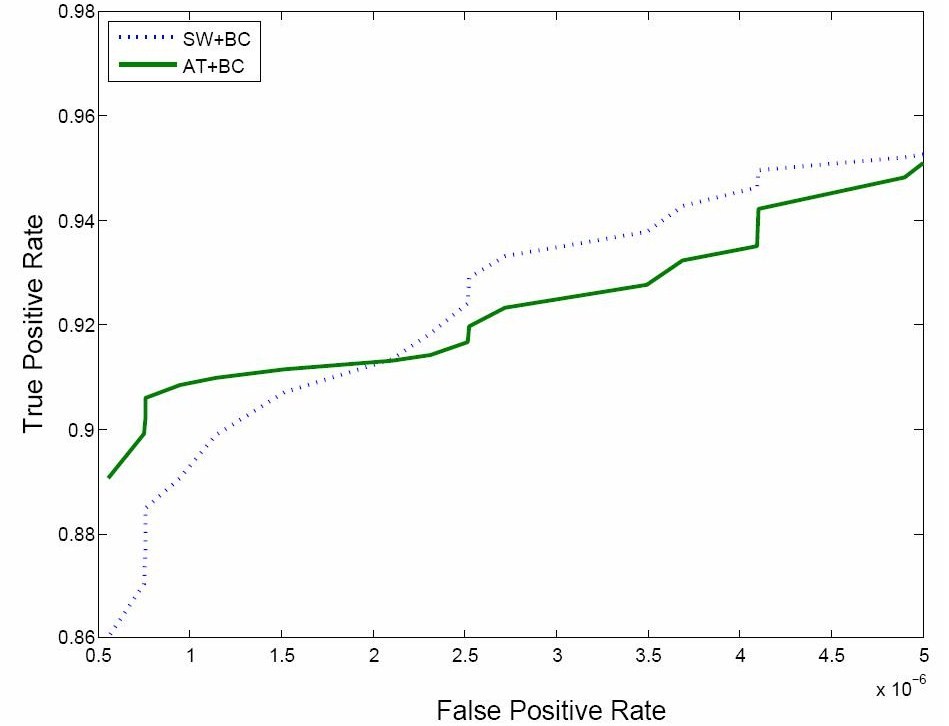}}
\subfigure[ ]{\label{fig:loc_time}
\includegraphics[width=0.3\textwidth]{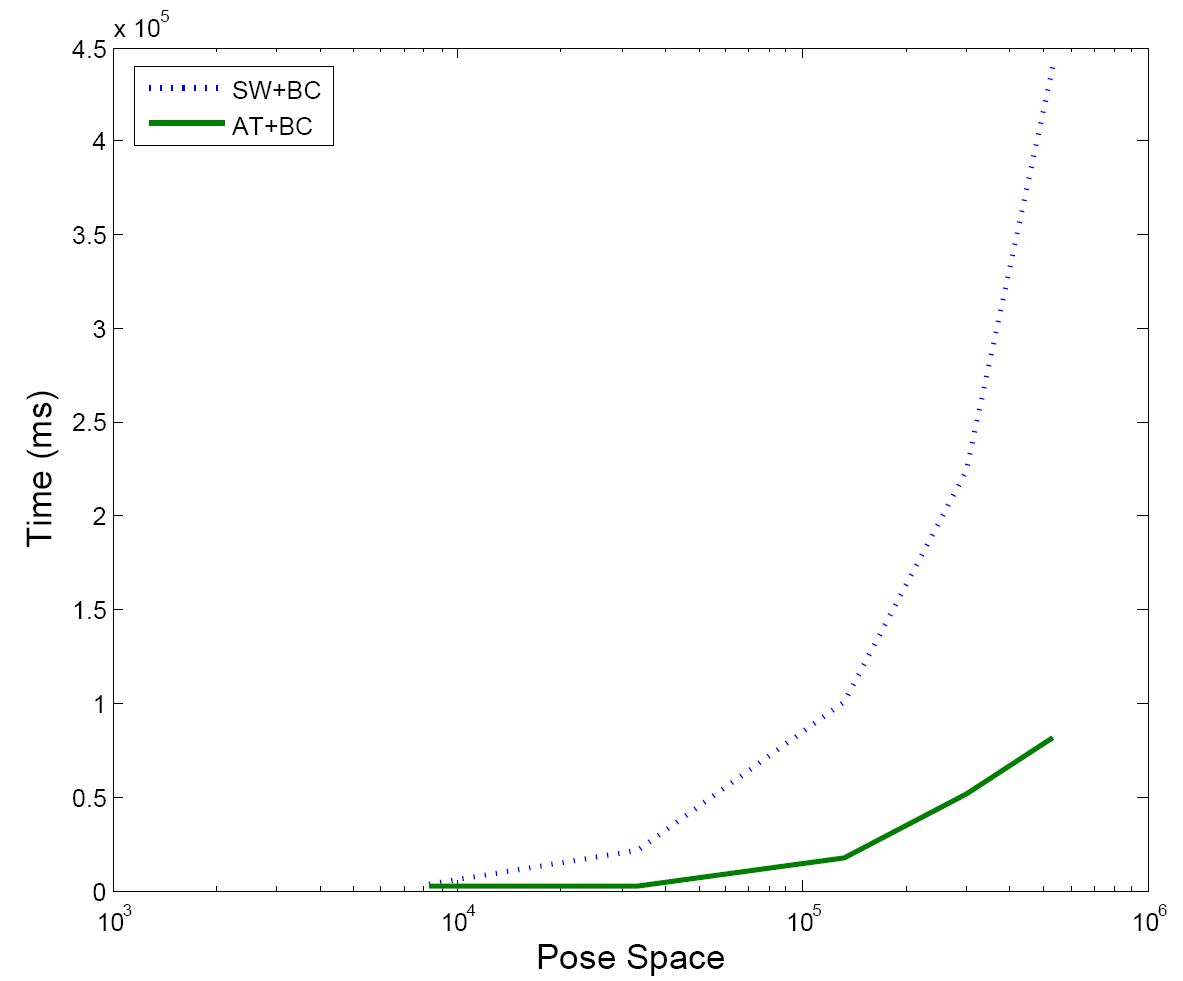}}
\subfigure[ ]{\label{fig:loc_eval}
\includegraphics[width=0.3\textwidth]{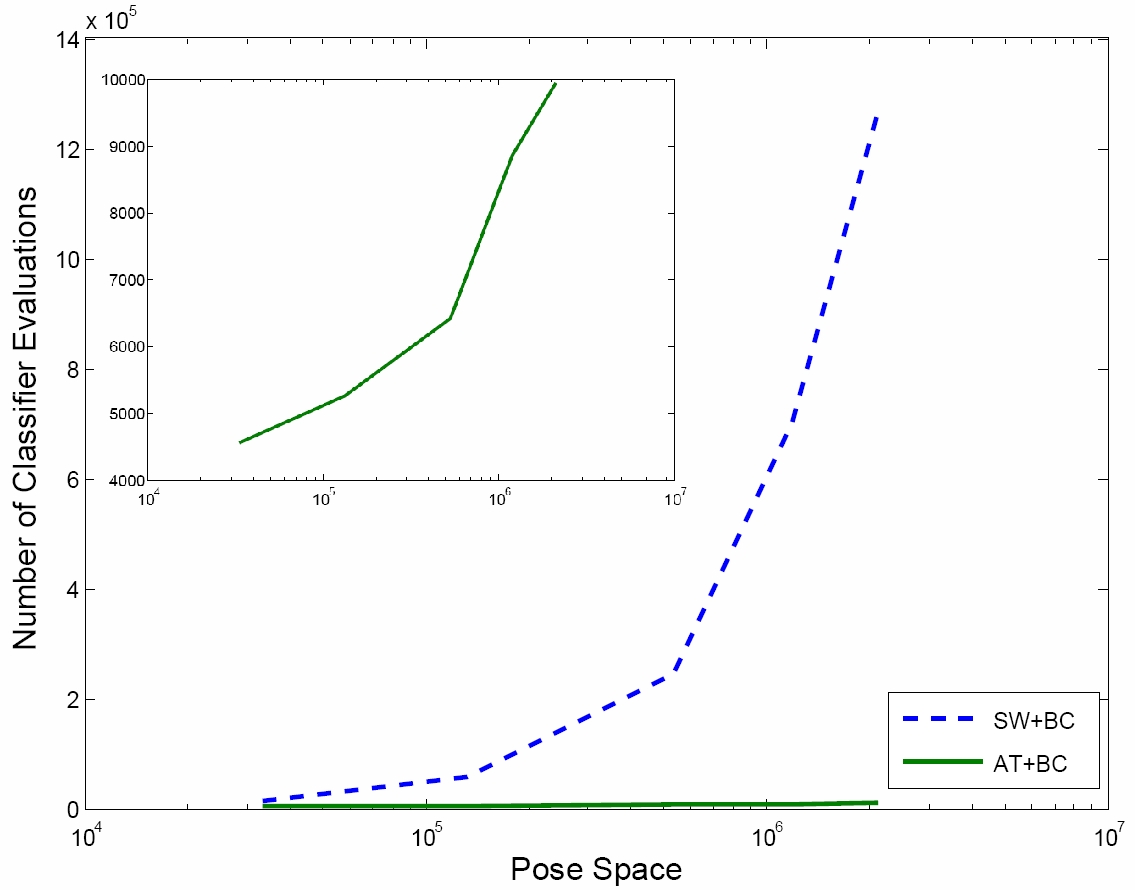}}
\caption{\small (a) ROC curve of both SW+BC and AT+BC to find a face
in the Caltech Frontal Face dataset. The performance of both methods
is approximately identical. (b) Average computation time with
varying pose space size. Note that image size is in logarithmic
scale. The AT algorithm performs in almost logarithmic time compared
to SW. (c) Average number of classifier evaluations when the pose
space increases. Additionally, a zoom of the AT performance is
provided. } \label{fig:localization}
\end{figure*}

\normalsize

\subsection{Offline Training}
\label{sec:exp_train}

We choose to model each $f^k_s(\cdot,i)$ with a Beta distribution
with parameters ($\alpha,\beta$). To do this, we randomly selected
$50$ images, from the Caltech Frontal Face Dataset \cite{weber99}.
Note that far fewer images are used for training here when compared
to other search methods (see \cite{Lampert08,yuan09}) which
typically use on the order of $10^3$ images to train their systems.
The estimation of the $f^k_s(\cdot,i)$ parameters is broken into two
parts.

We first estimate all the background densities. That is, for each
$k$ and $i$, we randomly select 100 $j$'s per image, such that the
face center is not in $\Lambda_{i,j}$. We then compute the tests
$X^k_{i,j}=x$, and use these to compute the parameters using maximum
likelihood estimation with 5000 datapoints.

To estimate the foreground densities, a similar procedure is used.
We describe the case $s=1$. For each $k$ and $i$, we randomly select
100 $j$'s in each image such that the face center is in
$\Lambda_{i,j}$. The parameters of $f^k_1(\cdot,i)$ are then
estimated from the tests $X^k_{i,j}=x$. As before 5000 datapoints
are used to estimate ($\alpha, \beta$). In order to estimate
$f^k_s(\cdot,i)$ for $1<s\leq 4$, we subsample the images and repeat
the same procedure (similar to \cite{fleuret01}). Additionally, the
$\int(f^k_s - f^k_m)^2$ term from equation (\ref{eq:phiFaceScale})
is then calculated by using a Monte Carlo approximation, and stored
in a look-up table.

\subsection{Single Face Localization}
\label{sec:exp_single}

\begin{figure}
\centering \subfigure[
]{\label{fig:loc_perf}\includegraphics[width=0.23\textwidth]{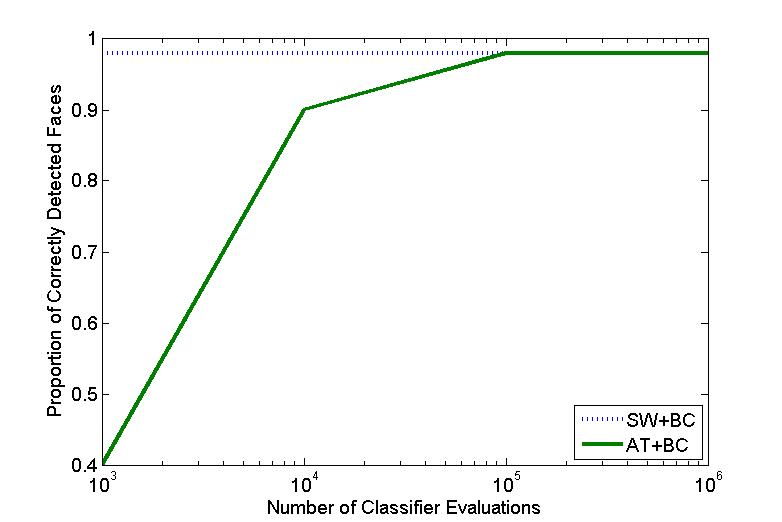}}
\subfigure[ ]{
\label{fig:loc_hist}\includegraphics[width=0.24\textwidth]{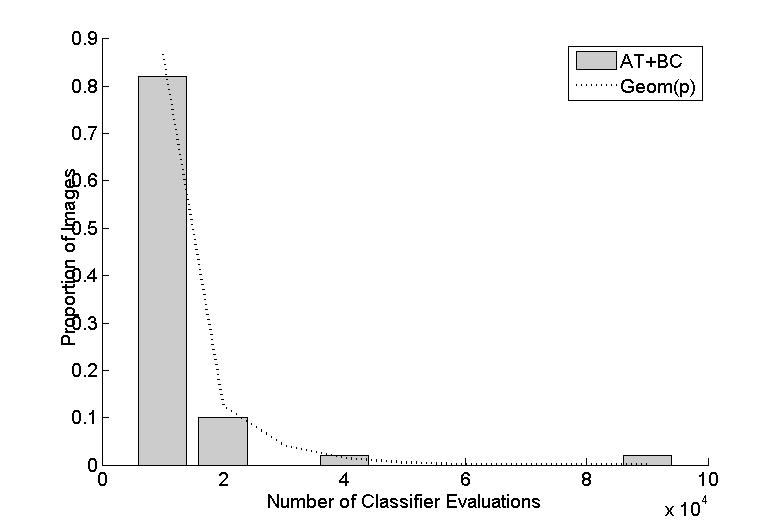}}
\subfigure[
]{\label{fig:loc_comp}\includegraphics[width=0.45\textwidth]{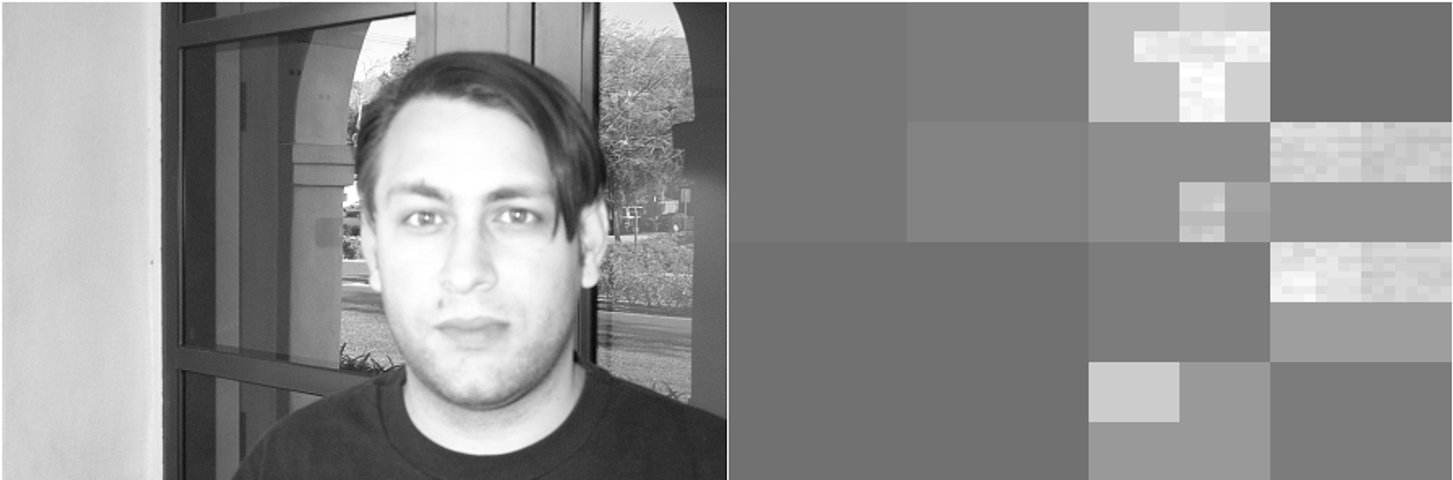}}
\caption{\small (a) The proportion of faces detected increases with the number of classifier evaluations:
$90\%$ of faces are
correctly detected with only $10^4$ evaluations and with $10^5$
classifier evaluations, the AT algorithm performs as well as SW, but
much faster. (b) Histogram of the number of classifier evaluations.
The dotted black line represents the point mass function of the
Geometric distribution with parameter $p = 1/9248$. (c) Face image
and associated computation image. This gray scaled image indicates
the number of times each pixel has been included in a queried
window. } \label{fig:classifier_single_face}
\end{figure}

We setup the AT algorithm with BCs (AT$+$BC) to run
until a face is found or until $5 \times 10^5$ classifier
evaluations have been performed (see figure
\ref{fig:classifier_single_face} for details on how this was
chosen). We compare this with a sliding window approach using the
identical BCs (SW$+$BC) and letting it run until a face
is found or until all poses have been observed. Note that both
(AT$+$BC) and (SW$+$BC) have the same pose space: all pixels and
face sizes (\eg pose space size $=896 \times 592 \times 13 =
6895616$). In order to avoid any unfair bias as to where faces may
be located, we randomly pick initial starting locations in the image
for (SW$+$BC), looping around the image in order to observe all the
poses. We report that (AT$+$BC) allows for exponential computational
gains over the sliding window approach while keeping similar
performance levels.

Figure \ref{fig:seq} shows a typical behavior of the AT algorithm on
a given image. In general, the order in which queries are posed is
complex and in some cases counter-intuitive - validating the need
for an \textit{online} search strategy.

In figure \ref{fig:localization}(a) we compare the accuracy of (AT$+$BC) and
(SW$+$BC) on the remaining unused 400 images of the dataset using a
ROC curve. We observe that generally (AT$+$BC) does not suffer much
from a loss in performance compared to the brute force sliding
window approach. Note that the difference between the two methods is
not significant.

To compare how much time (AT$+$BC) and (SW$+$BC) take to locate a
face depending on the size of the pose space, we randomly selected a
subset of 50 images from the testing set, subsampled these to have
images of sizes ($112 \times 74$, $ 224 \times 148 $, $ 448 \times
296$, $672 \times 444$, $896 \times 592 $). Figure
\ref{fig:localization}(b), shows the average time of both methods for each
image size. Note that the overhead of (AT$+$BC) - the time to
evaluate all queries tested, the update mechanism and the query
selection - is included in this plot (the additional time to compute
an integral image for oriented edges is not included as it is
negligible). As expected, we see that (SW$+$BC) is linear in the
number of poses. However, the total time (AT$+$BC) takes to complete
is significantly lower than (SW$+$BC) and even more so at large
image sizes. In fact, (AT$+$BC) remains almost logarithmic even as
the number of poses increases. This suggests that AT uses a form of
``Divide and Conquer'' search strategy. Note, that at image sizes
smaller than ($112 \times 74$), (AT$+$BC) is slower than (SW$+$BC)
due to the overhead.

Figure \ref{fig:localization}(c) shows the average number of classifier
evaluations both (AT$+$BC) and (SW$+$BC) perform, when changing the
image size. Notice that the difference between (AT$+$BC) and
(SW$+$BC) is even larger than the difference reported in figure
\ref{fig:localization}(b), and that the AT algorithm significantly reduces
the number of classifier evaluations. For the largest image size AT
requires 100 times fewer evaluations than SW.

In figure \ref{fig:classifier_single_face}(a) we show how the accuracy of (AT$+$BC)
is affected by the total number of classifier evaluations allowed.
The dotted line indicates the performance of (SW$+$BC) when the
entire pose space is observed. We see that after observing the
entire pose space ($O(10^6)$ evaluations), 98\% accuracy is
achieved. Performance results are shown when (AT$+$BC) is stopped
when either a face has been located or after ($10^3$, $10^4$,
$10^5$, $10^6$) classifier evaluations have be performed. After only
$10^4$ classifier evaluations nearly $90\%$ of detectable faces are
found. By $10^5$ evaluations AT performs at the same accuracy level
as SW. In general, we can see in figure \ref{fig:classifier_single_face}(b) that the
number of evaluations required is approximatively Geometric$(p = 10^{-4})$. Hence, on average $0.0014$
of the total pose space is evaluated by the classifier.

As in \cite{amit99}, figure \ref{fig:classifier_single_face}(c) shows a randomly
selected test image, and the corresponding computational image
associated (right). The computational image is a gray scale image,
which indicates the number of times each pixel has been included in
a queried window (all types of queries included). Darker regions
show areas where little computation has taken place, while white
regions shows important computation. As expected, we can see that
regions of the image which contain few features (left part of the
image) are not considered for much computation.

\normalsize

\section{Face Detection and Localization}
\label{sec:exp_multi}

We now test the AT algorithm in a much harder setting - a detection
and localization task. We do this by looking for faces in the
MIT+CMU dataset \cite{mitFaceSet}. This dataset contains 130 images,
of various sizes, where some images contain no faces, and others
contain an unknown number of faces. Face sizes range between 20
pixels to the width of images. As in the previous experiment, we
initialize the AT algorithm similarly to that in section
\ref{sec:model} and \ref{sec:experiment_loc}.

To find multiple face instances, we assume that at any point in
time, the remaining number of faces to be found in an image follows
a Poisson distribution with parameter $\lambda Q$, where $Q$ is the
number of pixels unobserved in the image, and $\lambda$ is a face
rate. We have chosen $\lambda = 10^{-4}$, corresponding to one face
per 100x100 pixel image on average and hence
$\pi_0(\bar{\Lambda}_{1,1})=e^{-\lambda Q}$. We then run the AT
algorithm until $\pi_t(\Lambda_{1,1}) < \epsilon = 10^{-5}$. When a
face is found: edges
from the detected face region are removed from the integral images
and the remaining poses are assigned uniform probability. The
algorithm is then restarted with the updated
$\pi_0(\bar{\Lambda}_{1,1})$.

\begin{figure}
\centering \subfigure[ ]{ \label{fig:det_roc}
\includegraphics[width=0.31\textwidth]{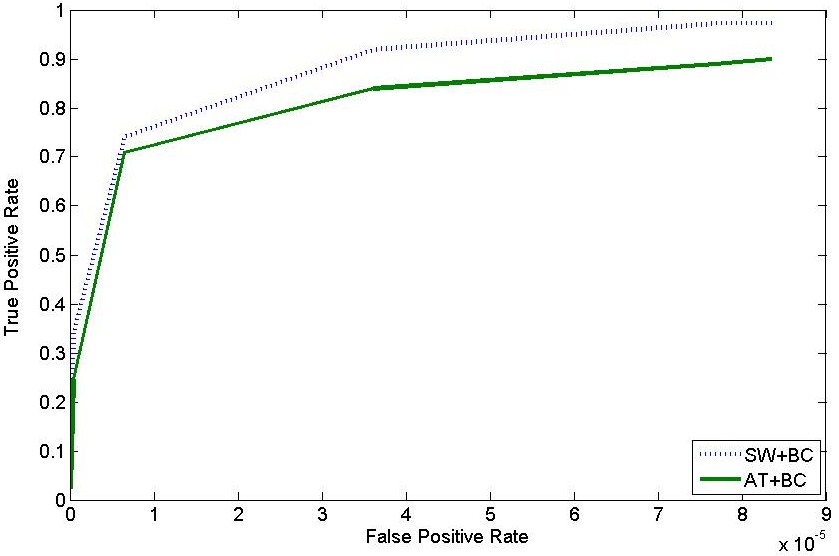}}
\subfigure[]{
\label{fig:det_eval}\includegraphics[width=0.31\textwidth]{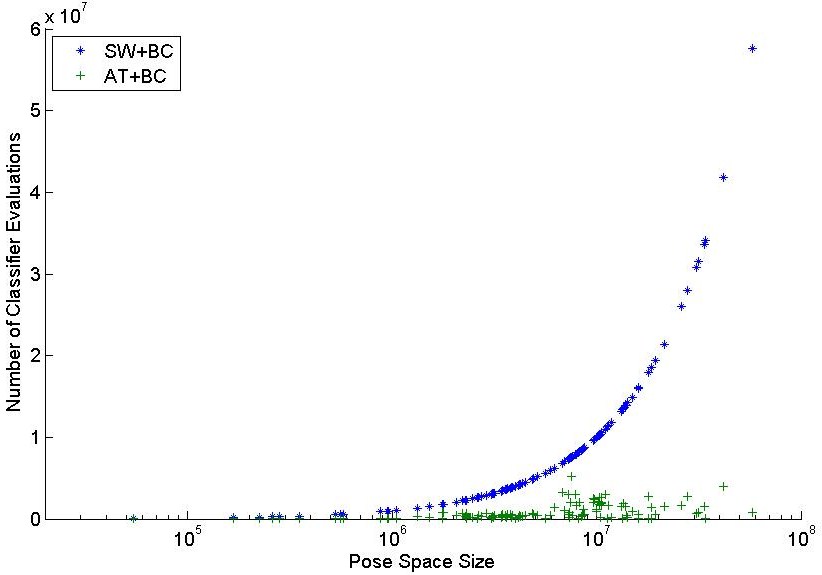}}
\subfigure[]{
\label{fig:det_time}\includegraphics[width=0.31\textwidth]{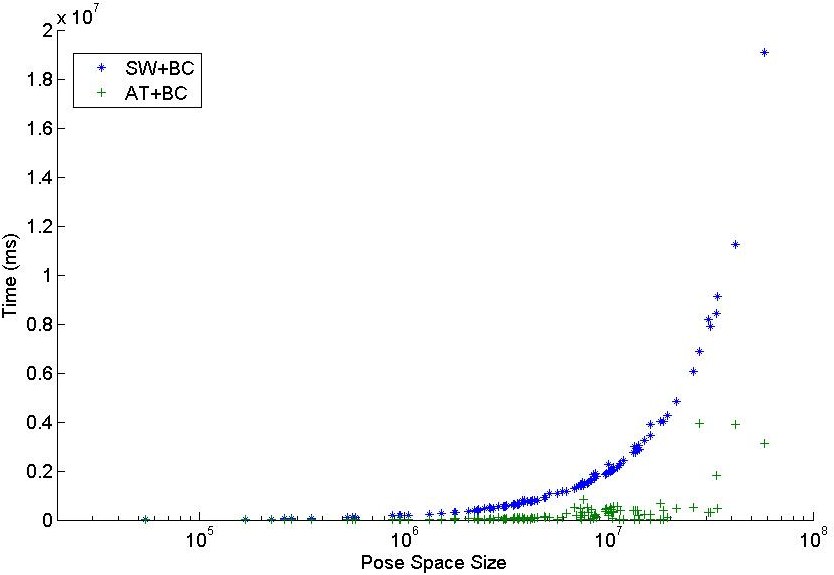}}
\caption{\small (a) ROC for both the sliding window and the Active
Testing approaches on the MIT+CMU frontal face dataset. The AT algorithm
achieves similar performance levels to the exhaustive search. (b)
Number of classifier evaluations for each image in the testset.
Clearly the AT approach does not suffer as much from the increase in
pose space. (c) Time performance for each image in the testset.}
\label{fig:multi_roc}
\end{figure}

Figure \ref{fig:multi_roc}(a) shows the ROC curve of both the (AT$+$BC) and (SW$+$BC) methods on the MIT+CMU dataset. In both cases no post-processing step was applied to these results (\ie No Non-Maximum suppression). First we
note that the MIT+CMU testset is much harder than the
Caltech Frontal Face set. In general, the performance
of the AT algorithm is comparable to the brute force
approach. There is, however, a slight performance
decrease in (AT$+$BC) when compared to the exhaustive search. That is, we notice that even though the classifier
used (BC) is not very good (when compared to state-of-
the-art classifiers), little accuracy loss is observed when
used in the AT framework.

From this experiment, (AT$+$BC) required $O(10^8)$ classifier
evaluations over the entire testset, while (SW$+$BC) required
$O(10^9)$ evaluations. Figure \ref{fig:multi_roc}(b), shows the number
of classifier evaluations required by both (AT$+$BC) and (SW$+$BC)
on each image. Generally, we see that AT is still able to
significantly reduce the total number of evaluations required even
though the number of faces in the images is apriori unknown. Figure
\ref{fig:multi_roc}(c), shows a similar result in terms of time. Again,
computational gains are of one order of magnitude over the entire
testset.

Notice in figures \ref{fig:multi_roc}(b) and \ref{fig:multi_roc}(c) that for
images of the same pose space size, the number of classifier
evaluations and time necessary for (AT$+$BC) to terminate vary. This
variance is due to the fact that (AT$+$BC) stops when the estimate
of having a face in the image is very low: $\pi_t(\Lambda_{1,1}) <
\epsilon = 10^{-5}$. Hence, in images which contain many face-like
features, the algorithm will need to visit many more locations to
see if faces are still present. This is precisely what is observed
in figures \ref{fig:multi_roc}(b) and \ref{fig:multi_roc}(c).

\normalsize

\section{Conclusion}
\label{sec:conclusion}

We have proposed an Active Testing framework in which one can
perform fast face detection and localization in images. In order to
find faces, we use a coarse-to-fine method, while sampling
subwindows which maximize information gain. This allows us to
quickly find the face pose by focusing on regions of interest, and
pruning large image regions. We show through a series of
experiments, that the active testing framework can be used to
significantly reduce the number of classifier evaluations when
searching for an object. Exponential speedup is observed when
detecting and locating faces compared to the traditional sliding
window approach (particularly on large image sizes), without
significant loss in performance levels, indicating that this
method is scalable to larger image sizes.

\section*{Acknowledgments}
Funding for this research was provided in part by NIH Grant 1 R01 EB 007969-01 and the Duncan Fund for the Advancement of Statistics Research, Award 08-19.

\bibliographystyle{plain}
\bibliography{active_testing_journalBib}

\end{document}